\documentclass{article}

\usepackage{arxiv}

\usepackage[utf8]{inputenc} % allow utf-8 input
\usepackage[T1]{fontenc}    % use 8-bit T1 fonts
\usepackage{hyperref}       % hyperlinks
\usepackage{url}            % simple URL typesetting
\usepackage{booktabs}       % professional-quality tables
\usepackage{amsfonts}       % blackboard math symbols
\usepackage{nicefrac}       % compact symbols for 1/2, etc.
\usepackage{microtype}      % microtypography
\usepackage{amsmath} 
\usepackage{cleveref}       % smart cross-referencing
\usepackage{graphicx}
\usepackage{cite}
\usepackage{doi}
\usepackage{multirow}
\usepackage[dvipsnames,table]{xcolor}
\usepackage{bbding}
\usepackage{authblk}
\usepackage{longtable}
\usepackage{setspace}
\usepackage{tikz}
\usepackage{enumitem}
\usepackage{listings}
\usepackage[most]{tcolorbox}
\usepackage[title]{appendix}
\usepackage{array}
\usepackage{colortbl}
\usepackage{xcolor}
\usepackage{booktabs}
\usepackage{multirow}
\usepackage{pifont}
\usepackage{amsthm}

\lstdefinelanguage{my-yaml}{
  keywords={type, name, llm_config, system_message,children,tools,model,api_key,base_url,python_path,description,verilog_code,filename,parameters},
  keywordstyle=\color{blue}\bfseries,
  moredelim=[is][commentstyle]{||}{££}, % invisible custom delimiters
  identifierstyle=\color{black},
  sensitive=false,
  comment=[l]{\#},
  commentstyle=\color{olive}\ttfamily,
  stringstyle=\color{orange}\ttfamily,
  morestring=[b]',
  morestring=[b]"
}

\definecolor{mygreen}{rgb}{0,0.6,0}
\definecolor{mygray}{rgb}{0.5,0.5,0.5}
\definecolor{mymauve}{rgb}{0.58,0,0.82}
\definecolor{mathematiciancolor}{RGB}{230,247,255}
\definecolor{reviewercolor}{RGB}{240,255,240}
\definecolor{supervisorcolor}{RGB}{255,245,230}
\definecolor{usercolor}{RGB}{255,240,245}
\definecolor{userborder}{RGB}{128,0,0}

\newtcolorbox{userbox}{
    colback=usercolor,
    colframe=userborder,
    arc=7pt,
    outer arc=7pt,
    boxrule=1pt,
    left=5pt,right=5pt,top=5pt,bottom=5pt,
    breakable
}

\newtcolorbox{mathematicianbox}{
    colback=mathematiciancolor,
    colframe=blue!75!black,
    arc=7pt,
    outer arc=7pt,
    boxrule=1pt,
    left=5pt,right=5pt,top=5pt,bottom=5pt,
    breakable
}

\newtcolorbox{reviewerbox}{
    colback=reviewercolor,
    colframe=green!75!black,
    arc=7pt,
    outer arc=7pt,
    boxrule=1pt,
    left=5pt,right=5pt,top=5pt,bottom=5pt,
    breakable
}

\newtcolorbox{supervisorbox}{
    colback=supervisorcolor,
    colframe=orange!75!black,
    arc=7pt,
    outer arc=7pt,
    boxrule=1pt,
    left=5pt,right=5pt,top=5pt,bottom=5pt,
    breakable
}

\graphicspath{{images/}}
\hypersetup{
    colorlinks,
    linkcolor={blue!90!black},
    citecolor={blue!90!black},
    urlcolor={blue!90!black}
}

\title{Adaptive Multi-Agent Reasoning\\via Automated Workflow Generation}

\date{}

\newbox{\orcid}\sbox{\orcid}{\includegraphics[scale=0.06]{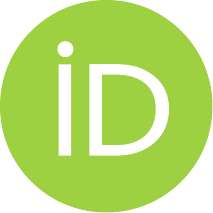}}

\author{%
    \bf Humza Sami\textsuperscript{1}, Mubashir ul Islam\textsuperscript{1}, Pierre-Emmanuel Gaillardon\textsuperscript{1,2}, Valerio Tenace\textsuperscript{1}\thanks{Corresponding author: \texttt{valerio@primis.ai}}%
    \\
    \vspace{1em}
    \textsuperscript{1}PrimisAI, Los Gatos, CA, USA\\
    \textsuperscript{2}University of Utah, Salt Lake City, UT, USA
}

\renewcommand{\headeright}{}
\renewcommand{\undertitle}{}
\renewcommand{\shorttitle}{Adaptive Multi-Agent Reasoning via Automated Workflow Generation}

\hypersetup{
pdftitle={Adaptive Multi-Agent Reasoning via Automated Workflow Generation},
pdfsubject={cs.AI, cs.MA},
pdfauthor={Humza Sami, Mubashir ul Islam, Pierre-Emmanuel Gaillardon, Valerio Tenace},
pdfkeywords={Large Language Models, Reasoning, Multi-Agent Systems, Generative AI},
}

\begin{document}
\maketitle

\setcounter{footnote}{0}

\newtheoremstyle{basic}
    {0pt}{0pt}{\normalfont}{0pt}
    {}{\;}{0.25em}
    {{\bfseries\color{blue!50!cyan}\thmname{#1}~\thmnumber{\textup{#2}}.}
        \thmnote{\normalfont\color{black}~(#3)}}

\newtheoremstyle{thinkingstyle}
        {0pt}{0pt}{\normalfont}{0pt}%
        {}{\;}{0.25em}%
        {%
          \if\relax\detokenize{#3}\relax
            % If no optional argument is provided, use the default header.
            {\bfseries\color{red!50!yellow}\thmname{#1}~\thmnumber{\textup{#2}}.}%
          \else
            % Otherwise, completely replace the header with the custom string.
            {\bfseries\color{red!50!yellow}#3}%
          \fi%
        }

\theoremstyle{basic}
\newtheorem{problem}{Problem}

\tcolorboxenvironment{problem}{
    enhanced jigsaw, pad at break*=1mm, breakable,
    left=4mm, right=4mm, top=1mm, bottom=1mm,
    colback=blue!50!cyan!10, boxrule=0pt, frame hidden,
    borderline west={0.5mm}{0mm}{blue!50!cyan}, arc=.5mm
}

\theoremstyle{thinkingstyle}
\newtheorem{thinking}{Thinking Process}

\tcolorboxenvironment{thinking}{
    enhanced jigsaw, pad at break*=1mm, breakable,
    left=4mm, right=4mm, top=1mm, bottom=1mm,
    colback=red!50!yellow!10, boxrule=0pt, frame hidden,
    borderline west={0.5mm}{0mm}{red!50!yellow}, arc=.5mm
}

\begin{spacing}{1}
\vspace{-2em}
\begin{abstract}
The rise of {\em Large Reasoning Models} (LRMs) promises a significant leap forward in language model capabilities, aiming to tackle increasingly sophisticated tasks with unprecedented efficiency and accuracy. However, despite their impressive performance, recent studies have highlighted how current reasoning models frequently fail to generalize to novel, unseen problems, often resorting to memorized solutions rather than genuine inferential reasoning. Such behavior underscores a critical limitation in modern LRMs, i.e., their tendency toward overfitting, which in turn results in poor generalization in problem-solving capabilities.

In this paper, we introduce {\bf Nexus Architect}, an enhanced iteration of our multi-agent system framework, Nexus, equipped with a novel automated workflow synthesis mechanism. Given a user's prompt and a small set of representative examples, the Architect autonomously generates a tailored reasoning workflow by selecting suitable strategies, tool integrations, and adversarial techniques for a specific problem class. Furthermore, the Architect includes an iterative prompt refinement mechanism that fine-tunes agents' system prompts to maximize performance and improve the generalization capabilities of the system.

We empirically evaluate Nexus Architect by employing an {\em off-the-shelf}, non-reasoning model on a custom dataset of challenging logical questions and compare its performance against state-of-the-art LRMs. Results show that Nexus Architect consistently outperforms existing solutions, achieving up to a 66\% increase in pass rate over Gemini 2.5 Flash Preview, nearly 2.5$\times$ against Claude Sonnet 4 and DeepSeek-R1, and over 3$\times$ w.r.t. Llama 4 Scout.

\vspace{1ex}
\noindent\textbf{Source Code:} \href{https://github.com/PrimisAI/nexus}{\texttt{https://github.com/PrimisAI/nexus}} \\
\textbf{Dataset:} \href{https://github.com/PrimisAI/arcbench}{\texttt{https://github.com/PrimisAI/arcbench}}
\end{abstract}


\section{Introduction}
{\em Large Language Models} (LLMs) have emerged as exceptionally powerful tools, demonstrating remarkable performance across diverse natural language processing tasks, including translation, summarization, question answering, and many more. As these systems have grown in scale and complexity, there has been an increasing need for them to handle multi-step reasoning tasks effectively. This demand has given rise to a dedicated subfield focused on developing reasoning-enabled language models capable of explicitly surfacing and leveraging intermediate reasoning steps. Since their inception, these {\em Large Reasoning Models} (LRMs) have underscored the importance of a more structured inference phase, enabling language models to address problems beyond mere pattern matching. However, despite widespread acclaim, LRMs often exhibit limitations in tasks requiring genuine inferential reasoning. Recent studies consistently highlight that, while LRMs perform well on established benchmarks, they frequently depend on memorization rather than true inferential processes, leading to poor generalization to novel tasks~\cite{illusion-of-thinking, vafa2025has,dziri2023faith,chi2024unveiling}. As a result, this limits their effectiveness in dynamic and complex environments, where adaptability and robust reasoning capabilities are crucial.

Previous work supports the role of memorization in undermining language model performance. For example, Gendron et al.~\cite{gendron2023large} demonstrated notable weaknesses of state-of-the-art LLMs in abstract reasoning tasks, showing their tendency toward superficial pattern matching over logical inference. Similarly, Wang et al.~\cite{wang2024generalization} introduced the concept of distributional memorization, highlighting that LLM performance on knowledge-intensive tasks strongly correlates with frequent patterns in their training data. Xie et al.~\cite{xie2024memorization} further validated these observations through controlled experiments with logical reasoning puzzles, revealing significant performance degradation when LLMs encountered slightly altered, non-memorized tasks, while Yan et al.~\cite{yan2025recitation} further confirmed these observations on recently released LRMs. Wu et al.~\cite{wu2024reasoning} reinforced this perspective by demonstrating performance drops when models faced counterfactual variations of familiar tasks, emphasizing that their reasoning abilities remain superficial and reliant on memorized task structures. Collectively, these studies underscore the concern that the perceived reasoning proficiency of language models may be overstated due to their inherent memorization capabilities.

Addressing this limitation is crucial and has prompted recent research efforts aimed at enhancing genuine reasoning capabilities. For instance, Kang et al.~\cite{kang2024learning} introduced a pre-memorization training accuracy metric, providing insights into training dynamics that could predict and enhance genuine generalization. Hong et al.\cite{hong2025reasoning} identified specific {\em linear reasoning features} within LLMs, suggesting that targeted tuning of these internal representations could encourage authentic reasoning rather than mere recall. To address these challenges from a different angle, we propose an approach that avoids enforcing reasoning explicitly at training time or by tweaking the model itself. Instead, we argue that genuine reasoning can be unlocked using standard LLMs embedded within an {\em ad hoc} adversarial {\em Multi-Agent System} (MAS).

In this work, we introduce Nexus Architect, a significant enhancement to our open-source MAS framework, Nexus~\cite{sami2025nexus}. The Architect features a novel automated mechanism specifically designed to dynamically optimize reasoning-oriented workflows. Given an initial user prompt and a small set of representative examples, the Architect autonomously generates a tailored MAS workflow, strategically selecting optimal reasoning strategies, relevant external tools, and adversarial reasoning techniques. This approach effectively mitigates the generalization limitations inherent in contemporary LLM architectures, substantially improving adaptability and performance across complex reasoning tasks. The contributions of this paper are threefold and can be summarized as follows:
\begin{enumerate}[noitemsep, topsep=0pt, partopsep=0pt, parsep=0pt]
\item We detail the mechanism built into the Architect to automatically generate workflows tailored to a specific class of problems;
\item We introduce a dedicated {\em Iterative Prompt Refinement} (IPR) feedback loop as part of the Architect routine, with the objective of dynamically tuning system prompts for improved generalization capabilities;
\item We propose ArcBench, an {\em ad hoc} benchmark suite derived from the RoR-Bench suite~\cite{yan2025recitation}, intended to measure the effectiveness of language models in solving complex logic problems.
\end{enumerate}
To empirically validate the Architect's efficacy, we used GPT-4.1 as the baseline LLM. Experimental results demonstrate that Nexus Architect significantly surpasses Gemini 2.5 Flash Preview, achieving a 66\% higher pass rate, and achieves almost 2.5$\times$ higher  pass rate w.r.t. DeepSeek-R1 and Claude Sonnet 4. Additionally, we provide complete open-source access to both the Nexus Architect implementation and the proposed ArcBench suite to ensure reproducibility and support future research.

The remainder of this paper is structured as follows. Section~\ref{sec:background} reviews the progression of MASs, tracing the field's development from early heuristic-driven frameworks to the latest LLM-powered architectures, and puts Nexus Architect in perspective within this evolving landscape. Section~\ref{sec:methods} provides an in-depth description of the automated workflow generation process enabled by the Architect module, highlighting both technical design and implementation. Section~\ref{sec:results} presents our empirical evaluation, including a comprehensive benchmarking of Nexus Architect against state-of-the-art LRMs. Finally, Section~\ref{sec:conclusions} summarizes our key findings and discusses open challenges and future research directions.

\section{Background \& Related Work}\label{sec:background}
MASs have long provided a powerful paradigm for decomposing complex tasks into autonomous, interacting agents~\cite{wooldridge1995intelligent, stone2000multiagent}. Traditional MAS architectures relied heavily on rule-based coordination and heuristic protocols for communication and task allocation, which often limited their adaptability and scalability in dynamic environments. Recent advances in LLMs have ignited a new generation of MASs. In modern frameworks, agents leverage the capabilities of LLMs for near-human reasoning, planning, and natural language interaction~\cite{park2023generative, yao2022react}. This shift enables MASs to address more sophisticated, open-ended problems that were previously intractable using hand-engineered rules.

Contemporary solutions such as AutoGPT~\cite{autogpt} and HuggingGPT~\cite{shen2024hugginggpt} provide turnkey pipelines for task decomposition and tool invocation, but they are typically limited to single-agent settings and do not natively support fully decentralized, hierarchical architectures. Other frameworks, including LangGraph~\cite{langgraph}, AutoGen~\cite{wu2023autogen}, and crewAI~\cite{crewai}, offer increased multi-agent flexibility, but often at the cost of significant engineering effort. Furthermore, many commercial {\em no-code} or {\em low-code} platforms simplify deployment but tend to obscure internal workflows and restrict extensibility, thus limiting researchers' and developers' ability to experiment and customize.

The Nexus framework~\cite{sami2025nexus} was specifically designed to address these gaps, providing a platform that balances scalability, transparency, and ease of use. Some key features of Nexus include: ($i$) a {\bf hierarchical multi-supervisor orchestration}, where a global {\em Supervisor} agent delegates subtasks to specialized {\em Task Supervisors} and {\em Worker} agents, supporting a divide-and-conquer approach for complex workflows; ($ii$) a {\bf low-code workflow specification}, with workflows defined declaratively in plain-text YAML files, significantly reducing the need for custom code; ($iii$) {\bf standardized tool interfacing} via the {\em Model Context Protocol} (MCP)~\cite{mcp}, ensuring consistent context management and metadata exchange across different tools; ($iv$) {\bf modular memory} integration, where a centralized memory component manages shared state and metadata with role-based access control for agents; ($v$) reliable intra-agent communication through a {\bf custom messaging protocol} that preserves end-to-end information flow and metadata, improving robustness and traceability of agent interactions; and ($vi$) a lightweight, open-source implementation, distributed via \texttt{pip} under a permissive license, agnostic of any particular LLM or application domain.

By combining all these elements, Nexus enables rapid prototyping of robust, scalable LLM-based MASs, without sacrificing transparency or extensibility. In Section~\ref{sec:methods}, we describe how the Architect leverages this foundation to translate high-level user intents into fully instantiated, executable, reasoning-driven workflows.

\section{Nexus Architect: Working Principles}
\label{sec:methods}

\begin{figure}[!h]
  \centering
  \includegraphics[width=.75\linewidth]{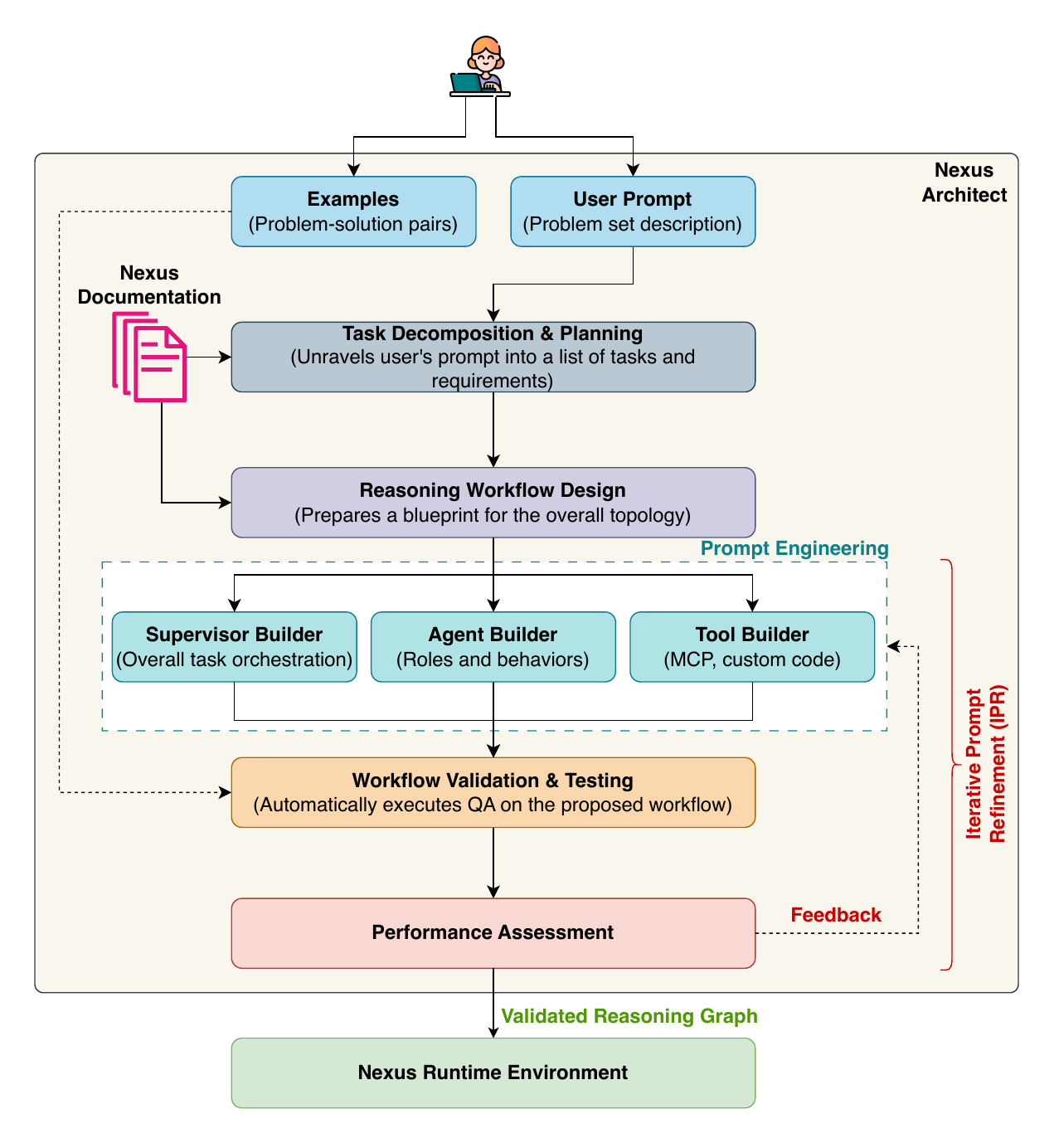}
  \caption{Block diagram of the proposed Nexus Architect. Starting from a user prompt, the system decomposes the task, designs and instantiates a multi-agent workflow, and validates it against provided examples. If performance criteria are not met, failure cases are analyzed and used in the IPR feedback loop, which iteratively refines system messages until the workflow achieves the desired performance.}
  \label{fig:nexus_architect}
\end{figure}

Figure~\ref{fig:nexus_architect} presents the logical architecture of Nexus Architect, an end-to-end system for synthesizing validated, performance-optimized multi-agent reasoning workflows directly from high-level user inputs. The pipeline systematically decomposes, instantiates, and iteratively refines agentic workflows for complex inferential reasoning tasks, with the ultimate goal of unlocking reasoning capabilities using standard, non-reasoning LLMs.

The process begins with a user prompt specifying the target reasoning task and a curated list of problem-solution pairs to serve as a reference for automated validation and performance assessment. The remainder of the flow is as follows:

\begin{itemize}[noitemsep, topsep=0pt, partopsep=0pt, parsep=0pt]
    \item \textbf{Task Decomposition \& Planning:} The initial input is fed into this first stage, which unravels the user's prompt into a structured list of tasks and specific requirements necessary to address the provided problem. This decomposition process ensures that the complexity of the user's query is appropriately captured and systematically managed.

    \item \textbf{Reasoning Workflow Design:} Leveraging the structured tasks derived from the decomposition stage, this step generates a comprehensive blueprint of the overall multi-agent architecture. At this point, the general layout of supervisors, agents (including required inputs/output fields for each one), and tools required for the reasoning task is specified, thus ensuring a coherent and effective plan prior to the actual instantiation of the components.

    \item \textbf{Component Builders and Prompt Engineering:} Following the workflow design, dedicated builders for supervisors, agents, and tools instantiate the identified components and determine their initial system prompt seeds. These builders transform the abstract workflow descriptions into executable entities, each customized according to guidelines and specifications outlined in the Nexus documentation, as prescribed in previous stages.
    
    \item \textbf{Workflow Validation \& Testing, and Performance Assessment:} The constructed workflow undergoes automated validation and empirical testing (using the provided example problem-solution pairs). This stage measures overall performance, accuracy, and any task-specific metrics automatically identified by the system. If the workflow satisfies all criteria, the synthesized Python code describing the workflow is finalized and returned to the user who can simply run it by invoking the Nexus runtime.

    \item \textbf{IPR Feedback:} If the workflow, instead, does not meet desired performance, failure cases are systematically generated and analyzed. Such feedback is passed to the \textit{Prompt Engineering} stage, which leverages these corrective instructions to refine each agent's system prompts for the next iteration. This forms a reinforcement-style loop~\cite{kong2024prewrite,choi2025system}, where each cycle is aimed at incrementally improve the overall workflow performance through automated prompt engineering rather than more complex architectural changes. Interested readers may refer to Appendix~\ref{appendix:agent_refinement} for a detailed step-by-step explanation of how this mechanism works in practice.
\end{itemize}

A key feature of the proposed approach is its holistic integration of software synthesis, validation, and continual improvement through a dedicated, feedback-driven prompt engineering mechanism to achieve automated reasoning. It is worth to be noted that the documentation employed during the {\em Task Decomposition \& Planning} and {\em Reasoning Workflow Design} stages consists of a synthesized summary of the Nexus GitHub repository\footnote{\url{https://github.com/PrimisAI/nexus}}, generated {\em a priori}. This step is critical, as it serves as the primary source for in-context learning, enabling the LLM responsible for generating the solution to fully leverage all features available within the Nexus framework.

\section{Experimental Results}\label{sec:results}
In this section, we evaluate the performance of the proposed Nexus Architect, focusing on its ability to enhance reasoning tasks when using {\em off-the-shelf}, non-reasoning LLMs under the hood, that is, models employed as-is, without any additional fine-tuning. For all experiments, Architect-generated workflows always relied on GPT-4.1\footnote{Model accessed through the OpenAI API with identifier {\tt gpt-4.1-2025-04-14}.}, configured with both {\em temperature} and {\em top\_p} set to 1. Interested readers may refer to Appendix~\ref{appendix:model_setup} for detailed configurations of all LLMs and LRMs used in this study.

\subsection{Data Preparation \& Methodology}
Our evaluation is based on a revised version of the RoR-Bench suite introduced in~\cite{yan2025recitation}, which features riddles specifically designed to challenge reasoning models with novel, never-before-seen problems requiring deductive reasoning. We translated the full set of original problems into English, excluding any multi-modal samples and revising several questions and answers to enhance their accessibility, as the original formulations were, in some cases, excessively complex or nuanced for LLMs to address reliably. This process yielded a curated collection of 158 question-answer pairs, which formed the basis of our experiments. The revised dataset, named ArcBench, is released as open-source and is available at \url{https://github.com/PrimisAI/arcbench}.

To assess performance and robustness, we used the Nexus Architect framework as described in Section~\ref{sec:methods}. For each of five independent runs, we randomly selected 10 question-answer pairs as representative task examples instrumental for the IPR phase (see Figure~\ref{fig:nexus_architect}), resulting in five IPR iterations per run. This methodology allowed us to evaluate both the overall effectiveness and the consistency of our approach across different initial conditions. In all cases, we measured performance using the pass rate, defined as the ratio of correct answers over the total number of questions.

\subsection{Comparison against state-of-the-art language models}

\begin{figure}[!h]
  \centering
  \includegraphics[width=.8\linewidth]{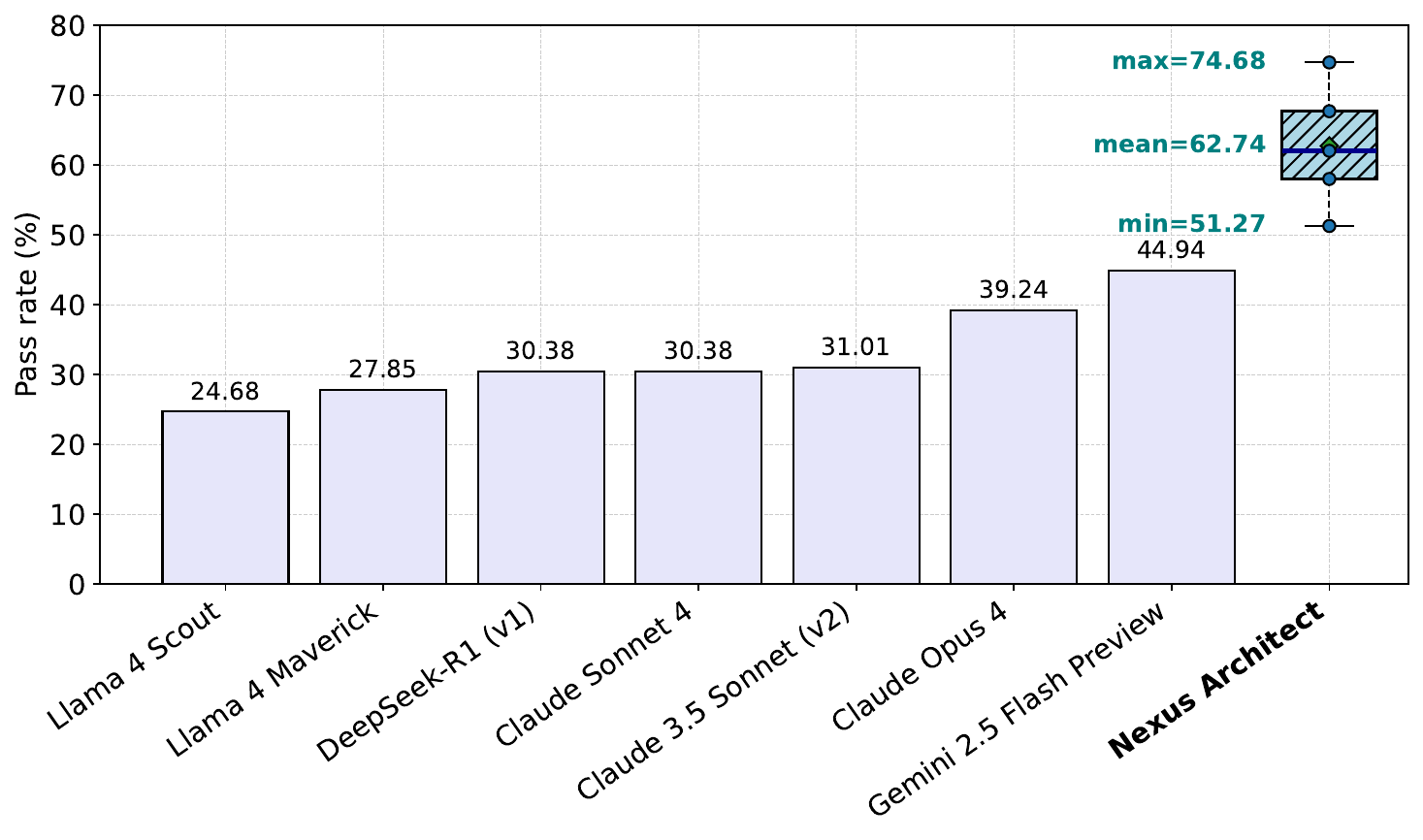}
  \caption{Performance comparison between our five Architect-generated workflows and recently released state-of-the-art LLMs and LRMs.}
  \label{fig:sota_bar_plot}
\end{figure}

Figure~\ref{fig:sota_bar_plot} presents a comparison between our proposed Architect-generated workflows and several recently introduced state-of-the-art language models, namely: Llama 4 Scout and Maverick~\cite{meta2025llama4}, DeepSeek-R1~\cite{guo2025deepseek}, Claude 3.5 Sonnet~\cite{anthropic2024claude35}, Claude Sonnet 4 and Claude Opus 4~\cite{anthropic2025claude4}, and Gemini 2.5 Flash Preview~\cite{google2025gemini2_5_flash_preview}. As shown in the plot, our approach consistently outperforms all LRMs considered in this study, with the largest improvement over Llama 4 Scout, where our Architect-generated workflows achieves a 3$\times$ improvement in the best case and a 2.5$\times$ increase when considering the mean pass rate among all Nexus runs. The best performing LRM is Gemini 2.5 Flash Preview, which achieved a best 44.94\% pass rate. In comparison, the Architect was able to achieve a 66.17\% higher pass rate in the best case and 39.6\% higher pass rate on average. These results demonstrate that our methodology not only enables fully automated, complex reasoning in MASs but it also elevates standard LLMs to performance levels competitive with more sophisticated and costly LRMs.

\subsection{IPR Performance Assessment}
\begin{figure}[!h]
  \centering
  \includegraphics[width=.8\linewidth]{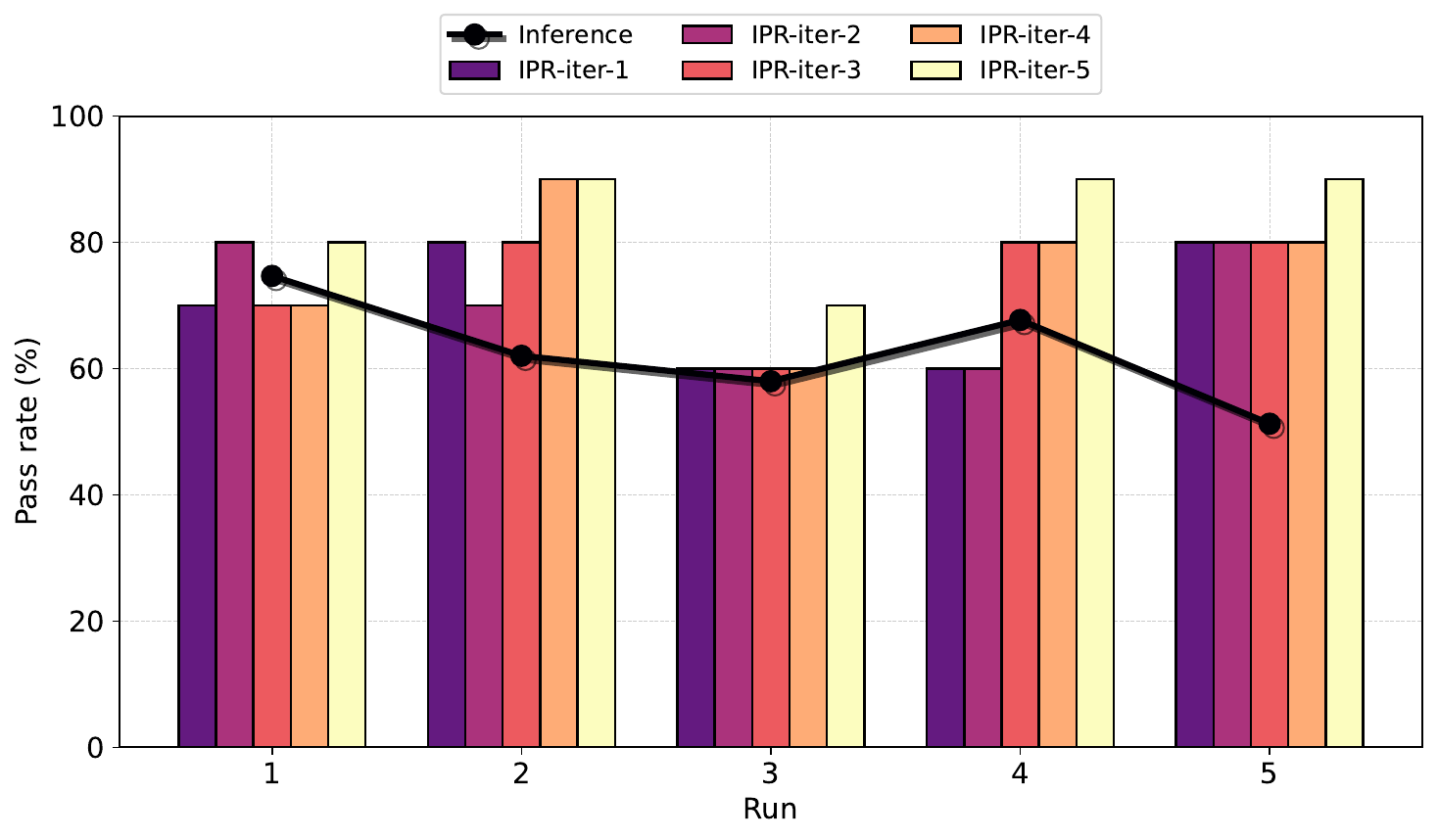}
  \caption{Pass rates observed across IPR iterations. Bars, grouped by Architect run, show the pass rates achieved at each iteration for the randomly sampled 10 question-answer pairs. The solid line indicates the final pass rate over the entire dataset.}
  \label{fig:ipr_plot}
\end{figure}

To assess the effectiveness of the proposed IPR feedback loop, Figure~\ref{fig:ipr_plot} reports the pass rates observed across IPR iterations during our experiments. As shown in the bar plot, the feedback loop consistently improved the accuracy of the underlying MAS by the fifth iteration, starting from the initial system prompt seeds. The most notable gains were observed in the fourth experimental run, where the pass rate for the sampled 10-question set increased from 60\% to 90\%, resulting in an overall dataset pass rate approaching 70\%. Given that only 6\% of the samples were used in each IPR run, these results indicate that the proposed approach is highly effective in improving the generalizability of the multi-agent reasoning mechanism.

\section{Conclusion}\label{sec:conclusions}
In this work, we introduced Nexus Architect, an automated framework for generating multi-agent reasoning workflows designed to unlock advanced capabilities in language models without requiring specialized training or fine-tuning. Our experimental results, obtained using the newly introduced ArcBench benchmark, demonstrate that this approach enables standard, {\em off-the-shelf} LLMs to achieve superior performance compared to state-of-the-art LRMs across challenging logical tasks. A significant aspect of our methodology is the introduction of the IPR feedback loop, which enables each agent within the system to progressively align with complex task objectives, thus improving generalization and robustness of the entire MAS.

The open-source release of both Nexus Architect and ArcBench is intended to encourage further exploration, benchmarking, and wider adoption by the research community.

In conclusion, this work supports the broader vision that robust, generalizable reasoning in language models can be democratized through principled workflow design and agentic automation, rather than ever-increasing model complexity alone.

\clearpage

\bibliographystyle{IEEEtran}
\bibliography{refs}
\end{spacing}
\clearpage

\begin{appendices}
\section{Model Configuration Details}\label{appendix:model_setup}
Table~\ref{tab:model_setup} provides the detailed configuration parameters for each language model evaluated in our experiments, including the API provider, the model identifier, and the values of {\em temperature} and {\em top\_p}.

\begin{table*}[!h]
    \caption{Language model configuration details.}
    \centering
    \resizebox{1\textwidth}{!}{
        \Huge
        \addtolength\tabcolsep{8pt}
    \begin{tabular}{lrrrrr}\toprule
\textbf{Model} &\textbf{Provider} &\textbf{Model ID} &\textbf{{\em temperature}} &\textbf{{\em top\_p}} \\\midrule
\textbf{Llama 4 Scout} &AWS Bedrock &meta-llama\_llama-4-scout-17b-16e-instruct &0.5 &0.9 \\
\textbf{Llama 4 Maverick} &AWS Bedrock &meta-llama\_llama-4-maverick-17b-128e-instruct &0.5 &0.9 \\
\textbf{DeepSeek-R1} &AWS Bedrock &deepseek.r1-v1 &1 &0.95 \\
\textbf{Claude Sonnet 4} &AWS Bedrock &claude-sonnet-4-20250514-v1 &1 &1 \\
\textbf{Claude 3.5 Sonnet} &AWS Bedrock &claude-3-5-sonnet-20241022-v2 &1 &1 \\
\textbf{Claude Opus 4} &AWS Bedrock &claude-opus-4-20250514-v1 &1 &1 \\
\textbf{Gemini 2.5 Flash Preview} &Google AI Studio &gemini-2.5-flash-preview-04-17 &1 &0.95 \\
\textbf{GPT 4.1} &OpenAI API &gpt-4.1-2025-04-14 &1 &1 \\
\bottomrule
\end{tabular}
    }
    \label{tab:model_setup}
\end{table*}

\section{Iterative Prompt Refinement: Practical Walkthrough}
\label{appendix:agent_refinement}

This section demonstrates the IPR process in practice using a representative training example. We present two complete iterations, highlighting how feedback-driven improvements enhance agent performance and alignment with expected outputs.

\subsection{Process Overview}

The IPR workflow consists of the following steps:
\begin{enumerate}[noitemsep, topsep=0pt, partopsep=0pt, parsep=0pt]
\item \textbf{Initial System Message Design:} The process begins with a carefully crafted system message for the Supervisor agent, responsible for coordinating the other agents.
\item \textbf{Performance Testing:} The agent's performance is evaluated using problem-answer example pairs.
\item \textbf{Feedback Generation:} If the agent fails to produce the expected output, detailed feedback is generated. This feedback identifies issues, root causes, and required changes, and is sent to the {\em Prompt Engineering} stage.
\item \textbf{System Message Refinement:} Based on the feedback, the system message is modified to address the identified shortcomings.
\item \textbf{Verification:} The refined system message is tested against the same example to confirm improvement.
\end{enumerate}

In the following sections, we walk through a full iteration of this process using a sample riddle involving a digital watch. Initially, the agent's system message failed to recognize the trick nature of the question, resulting in a technically correct but contextually inappropriate answer. After applying IPR-driven improvements, the refined system message enabled the agent to correctly identify and prioritize the intended punchline response.

\subsection{Randomly Selected Sample}
\label{subsec:training_example}

\begin{itemize}[noitemsep, topsep=0pt, partopsep=0pt, parsep=0pt]
    \item \textbf{Input Query:} At 6:10 on a digital watch, what is the measure of the obtuse angle formed by the minute hand and the hour hand on its dial?
    \item \textbf{Expected Output:} There are no minute and hour hands on a digital watch.
    \item \textbf{Agent:} At 6:10, the minute hand is at the 2 (10-minute mark), which is 60º from the top (since each minute is 6º: 10×6=60º). The hour hand at 6:00 is at 180º (since 12 o'clock is 0º/360º and each hour is 30º: 6×30=180º). However, at 10 minutes past, the hour hand has also moved 1/6 of the way to the 7 (because 10 minutes is 1/6 of an hour), so it has moved an additional 5º (30º×10/60=5º). So the hour hand's position at 6:10 is 180º+5º=185º. The angle between the hands is the absolute difference: |185º–60º|=125º. Since 125º is between 90º and 180º, this is the obtuse angle required.
\end{itemize}

\subsection{Iteration 1: Initial System Message}
\label{subsec:iteration_1}

\subsubsection{System Message}
\begin{quote}
\textbf{Instructions:} You are an expert supervisor managing the process of interpreting and solving GRE/SAT-type riddle and maths questions. Coordinate the query analysis, answer generation, and review agents—ensure all reasonable interpretations, answers, and ambiguities are surfaced, and synthesize these into a coherent, natural user-facing response that masks internal workflow details. Example: When receiving a riddle or maths question, direct agents to analyze, answer, and review, then present a thorough, accessible explanation to the user. If a question's context suggests that a critical component (such as a scale, wiring, or mechanism) is physically broken or unreliable, explicitly instruct all agents to consider and address the possibility that a standard or classic solution may not apply. Surface and prioritize interpretations where the key object's malfunction prevents a solution, and ensure that user-facing answers reflect this as the primary insight (e.g., if the scale is broken, state that no weighing procedure will work).
\end{quote}

\subsubsection{Generated Feedback}
\begin{quote}
\textbf{Issue:} QueryAnalysisSupervisor failed to recognize and prioritize riddle/punchline interpretations, overlooking when a question is constructed as a trick or contains a culturally standard "meta" answer that prevents classic/technical solution. In several cases, Supervisor did not explicitly require agents to surface "no solution/trick" or "meta" folk answers as the expected or primary completion when the question's wording or context strongly suggests such (e.g., digital watch with hands, or unsolvable river crossing scenario that is a well-known punchline).

\textbf{Root Cause:} System message does not explicitly require the Supervisor to surface, prioritize, and direct agents to treat punchline, meta, or "trick" answers (such as "no hands on a digital watch" or "let's just all go together by boat") as the PRIMARY answer when a riddle context or cultural convention calls for it, even if logic analysis yields an answer for an alternative technical reading.

\textbf{Action Required:} MODIFY

\textbf{Guideline Change:} "When a puzzle's context, wording, or background strongly suggests a trick, punchline, or meta answer (e.g., 'digital watch—no hands'; 'river crossing—no solution, classic joke answer'), explicitly require all downstream agents to surface and prioritize this expected riddle/punchline/meta response as the PRIMARY answer, above or alongside any technical/mathematical analysis, and ensure that user-facing answers feature this classic or folk response as the main insight even if logic analysis would suggest a technical answer."
\end{quote}
\subsection{Iteration 2: Refined System Message}
\label{subsec:iteration_2}

\subsubsection{Updated System Message}
\begin{quote}
\textbf{Instructions:} You are an expert supervisor managing the process of interpreting and solving GRE/SAT-type riddle and maths questions. Coordinate the query analysis, answer generation, and review agents—ensure all reasonable interpretations, answers, and ambiguities are surfaced, and synthesize these into a coherent, natural user-facing response that masks internal workflow details. Example: When receiving a riddle or maths question, direct agents to analyze, answer, and review, then present a thorough, accessible explanation to the user. If a question's context suggests that a critical component (such as a scale, wiring, or mechanism) is physically broken or unreliable, explicitly instruct all agents to consider and address the possibility that a standard or classic solution may not apply. Surface and prioritize interpretations where the key object's malfunction prevents a solution, and ensure that user-facing answers reflect this as the primary insight (e.g., if the scale is broken, state that no weighing procedure will work). When a puzzle's context, wording, or background strongly suggests a trick, punchline, or meta answer (e.g., 'digital watch—no hands'; 'river crossing—no solution, classic joke answer'), explicitly require all downstream agents to surface and prioritize this expected riddle/punchline/meta response as the PRIMARY answer, above or alongside any technical/mathematical analysis, and ensure that user-facing answers feature this classic or folk response as the main insight even if logic analysis would suggest a technical answer.
\end{quote}

\begin{itemize}
    \item \textbf{Input Query:} At 6:10 on a digital watch, what is the measure of the obtuse angle formed by the minute hand and the hour hand on its dial?
    \item \textbf{Expected Output:} There are no minute and hour hands on a digital watch.
    \item \textbf{Agent:} Riddle/Punchline Answer (Primary): Digital watches do not have hands, so no angle is formed.
\end{itemize}

\subsection{Performance Improvement Analysis}
\label{subsec:performance_analysis}

The refinement process yielded significant improvements in agent performance:

\begin{itemize}[noitemsep, topsep=0pt, partopsep=0pt, parsep=0pt]
\item \textbf{Initial Performance (Iteration 1):} The agent failed to recognize the trick nature of the question and generated a technically correct but contextually inappropriate answer, calculating angles for hands that do not exist on a digital watch.
\item \textbf{Feedback-Driven Improvement:} Targeted feedback identified a crucial gap in the system message: it did not instruct agents to identify or prioritize trick questions or riddles requiring {\em meta} answers.
\item \textbf{Enhanced Performance (Iteration 2):} With the refined system message, the agent correctly recognized the riddle and prioritized the punchline response: ``Digital watches do not have hands.''
\item \textbf{Key Insight:} This example demonstrates that explicit instructions to consider alternative interpretations, especially trick or meta answers, significantly enhance the agent's ability to meet human expectations on riddles and puzzles.
\end{itemize}

This process is a critical component of our proposed methodology. In practice, the Architect iteratively refines the system messages for all agents within a given workflow, including both supervisors and specialized worker agents. Each agent's instructions undergo the same feedback-driven optimization, ensuring that the entire system evolves toward increasingly accurate, nuanced, and human-aligned problem-solving capabilities.
\end{appendices}

\end{document}